\documentclass[10pt,twocolumn,letterpaper]{article}

\usepackage{iccv}
\usepackage{times}
\usepackage{epsfig}
\usepackage{graphicx}
\usepackage{amsmath}
\usepackage{amssymb}
\usepackage{booktabs}
\usepackage{colortbl}
\usepackage[dvipsnames,svgnames]{xcolor}
\usepackage{multirow}
\usepackage{xcolor}
\usepackage{subcaption}
\usepackage[accsupp]{axessibility}
\definecolor{mycolor}{RGB}{255, 105, 180}
\newcommand{\mycolor}[1]{\textcolor{mycolor}{#1}}

\newcommand{\myparagraph}[1]{\vspace{-10pt}\paragraph{#1}}

\usepackage[breaklinks=true,bookmarks=false,colorlinks=true]{hyperref}

\iccvfinalcopy 

\def\iccvPaperID{****} 

\ificcvfinal\fi

\begin{document}

\title{Learning to Upsample by Learning to Sample}

\author{Wenze Liu\\
\and
Hao Lu\thanks{Corresponding author}\\
\and
Hongtao Fu\\
\and
Zhiguo Cao\\
\and
School of Artificial Intelligence and Automation,\\
Huazhong University of Science and Technology, China\\
{\sf\small \{wzliu,hlu\}@hust.edu.cn}
}

\maketitle
\ificcvfinal\thispagestyle{empty}\fi

\begin{abstract} 
   We present DySample, an ultra-lightweight and effective dynamic upsampler. While impressive performance gains have been witnessed from recent kernel-based dynamic upsamplers such as CARAFE, FADE, and SAPA, they introduce much workload, mostly due to the time-consuming dynamic convolution and the additional sub-network used to generate dynamic kernels. Further, the need for high-res feature guidance of FADE and SAPA somehow limits their application scenarios. To address these concerns, we bypass dynamic convolution and formulate upsampling from the perspective of point sampling, which is more resource-efficient and can be easily implemented with the standard built-in function in PyTorch. We first showcase a naive design, and then demonstrate how to strengthen its upsampling behavior step by step towards our new upsampler, DySample. Compared with former kernel-based dynamic upsamplers, DySample requires no customized CUDA package and has much fewer parameters, FLOPs, GPU memory, and latency. Besides the light-weight characteristics, DySample outperforms other upsamplers across five dense prediction tasks, including semantic segmentation, object detection, instance segmentation, panoptic segmentation, and monocular depth estimation. Code is available at \url{https://github.com/tiny-smart/dysample}.
\end{abstract}

\section{Introduction}
\begin{figure}[!t]
	\centering
	\includegraphics[width=\linewidth]{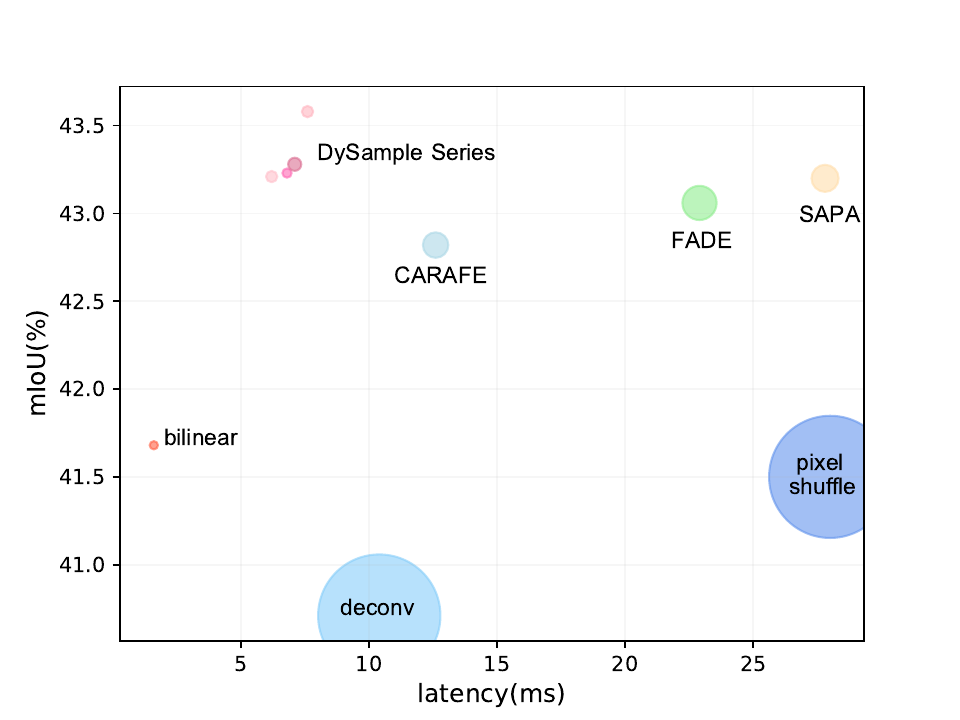}
	\caption{\textbf{Comparison of performance, inference speed, and GFLOPs of different upsamplers.} The circle size indicates the GFLOPs cost. The inference time is tested by $\times2$ upsampling a feature map of size $256\times120\times120$. The mIoU performance and additional GFLOPs are tested with SegFormer-B1~\cite{xie2021segformer} on the ADE20K data set~\cite{zhou2017scene}.
    }
	\label{fig:comparison}
\end{figure}

Feature upsampling is a crucial ingredient in dense prediction models for gradually recovering the feature resolution. The most commonly used upsamplers are nearest neighbor (NN) and bilinear interpolation, which follows fixed rules to interpolate upsampled values. To increase flexibility, learnable upsamplers are introduced in some specific tasks, \eg, deconvolution in instance segmentation~\cite{he2017mask} and pixel shuffle~\cite{shi2016real} in image super-resolution~\cite{mei2021image,haris2018deep,liang2021swinir}. However, they either suffer from checkerboard artifacts~\cite{odena2016deconvolution} or seem not friendly to high-level tasks. With the popularity of dynamic networks~\cite{jia2016dynamic}, some dynamic upsamplers have shown great potential on several tasks. CARAFE~\cite{jiaqi2019carafe} generates content-aware upsampling kernels to upsample the feature by dynamic convolution. The following work FADE~\cite{lu2022fade} and SAPA~\cite{lu2022sapa} propose to combine both the high-res guiding feature and the low-res input feature to generate dynamic kernels, such that the upsampling process could be guided by the higher-res structure. These dynamic upsamplers are often of complicated structures, require customized CUDA implementation, and cost much more inference time than bilinear interpolation. Particularly for FADE and SAPA, the higher-res guiding feature introduces even more computational workload and narrows their application scenarios (higher-res features must be available). Different from the early plain network~\cite{2015Fully}, multi-scale features are often used in modern architectures; therefore the higher-res feature as an input into upsamplers may not be necessary. For example in Feature Pyramid Network (FPN)~\cite{lin2017feature}, the higher-res feature would add into the low-res feature after upsampling. As a result, we believe that a well-designed single-input dynamic upsampler would be sufficient.

Considering the heavy workload introduced by dynamic convolution, we bypass the kernel-based paradigm and return to the essence of upsampling, \ie, point sampling, to reformulate the upsampling process. Specifically, we hypothesize that the input feature is interpolated to a continuous one with bilinear interpolation, and content-aware sampling points are generated to re-sample the continuous map. From this perspective, we first present a simple design, where point-wise offsets are generated by linear projection and used to re-sample point values with the $\tt{grid\_sample}$ function in PyTorch. Then we showcase how to improve it with step-by-step tweaks 
by i) controlling the initial sampling position, ii) adjusting the moving scope of the offsets, and iii) dividing the upsampling process into several independent groups, and obtain our new upsampler, DySample. At 
each step, 
we explain why the tweak is required and conduct experiments to verify the performance gain.

Compared with other dynamic upsamplers, DySample i) does not need high-res guiding features as input nor ii) any extra CUDA packages other than PyTorch, and particularly, iii) 
has much less inference latency, memory footprint, FLOPs, and number of parameters, as shown in Fig.~\ref{fig:comparison} and Fig.~\ref{fig:complexity}. For example, on semantic segmentation with MaskFormer-SwinB~\cite{cheng2021per} as the baseline, DySample 
invites $46\%$ more performance improvement than CARAFE, but 
requires only $3\%$ number of parameters and $20\%$ FLOPs of CARAFE. 
Thanks to the highly optimized PyTorch built-in function, the inference time of DySample also approaches to that of bilinear interpolation ($6.2$ ms vs.\ $1.6$ ms when upsampling a $256\times120\times120$ feature map).
Besides these appealing light-weight characteristics, DySample 
reports better performance compared with other upsamplers across five dense prediction tasks, including semantic segmentation, object detection, instance segmentation, panoptic segmentation, and monocular depth estimation. 

In a nutshell, we think DySample can safely replace NN/bilinear interpolation in existing dense prediction models, in light of not only effectiveness but also efficiency.

\section{Related Work}
We review dense prediction tasks, feature upsampling operators and dynamic sampling in deep learning.

\myparagraph{Dense Prediction Tasks.}
Dense prediction refers to a branch of tasks that require point-wise label prediction, such as semantic/instance/panoptic segmentation~\cite{badrinarayanan2017segnet,xiao2018unified,xie2021segformer,cheng2021per,cheng2021masked,he2017mask,fang2021instances,kirillov2019panopticfpn,li2021fully}, object detection~\cite{ren2015faster,cai2018cascade,lin2017focal,tian2019fcos}, and monocular depth estimation~\cite{wofk2019fastdepth,lee2019big,bhat2021adabins,li2022depthformer}. 
Different tasks often exhibit distinct characteristics and difficulties.
For example, it is hard to predict both smooth interior regions and sharp edges in semantic segmentation and also difficult to distinguish different objects in instance-aware tasks. In depth estimation, pixels with the same semantic meaning may have rather 
different depths, and vice versa. 
One often has to customize different architectures for different tasks.

Though model structure varies, upsampling operators are essential ingredients in dense prediction models. 
Since a backbone typically outputs multi-scale features, 
the low-res ones need to be upsampled to higher resolution. Therefore a light-weight, effective upsampler would benefit 
many dense prediction models. We will show our new upsampler design brings a consistent performance boost 
on SegFormer~\cite{xie2021segformer} and MaskFormer~\cite{cheng2021per} for semantic segmentaion, on Faster R-CNN~\cite{ren2015faster} for object detection, on Mask R-CNN~\cite{he2017mask} for instance segmentaion, on Panoptic FPN~\cite{kirillov2019panopticfpn} for panoptic segmentation, and on DepthFormer~\cite{li2022depthformer} for monocular depth estimation, while introducing negligible workload.

\myparagraph{Feature Upsampling.}
The commonly used feature upsamplers are NN and bilinear interpolation. They apply fixed rules to interpolate the low-res feature, ignoring the semantic meaning in the feature map. Max unpooling has been adopted in semantic segmentation by SegNet~\cite{badrinarayanan2017segnet} to preserve the edge information, but the introduction of noise and zero filling destroy the semantic consistency in smooth areas. Similar to convolution, some learnable upsamplers introduce 
learnable parameters in upsampling. For example, deconvolution upsamples features in a reverse 
fashion of convolution. 
Pixel Shuffle~\cite{shi2016real} uses convolution to increase the channel number ahead and then reshapes the feature map to 
increase the resolution.

Recently, some dynamic upsampling operators conduct content-aware upsampling. CARAFE~\cite{jiaqi2019carafe} uses a sub-network to generate content-aware dynamic convolution kernels to reassemble the input feature. FADE~\cite{lu2022fade} proposes to combine the high-res and low-res feature to generate dynamic kernels, for the sake of using the high-res structure. SAPA~\cite{lu2022sapa} further introduces the concept of point affiliation and computes similarity-aware kernels between high-res and low-res features. 
Being model plugins, these dynamic upsamplers 
increase more complexity than expected, especially for FADE and SAPA that require high-res feature input. 
Hence, our goal is to 
contribute a simple, fast, low-cost, and universal upsampler, while 
reserving the effectiveness
of dynamic upsampling.

\myparagraph{Dynamic Sampling.}
Upsampling is 
about modeling geometric information. A 
stream of work also models geometric information by dynamically sampling an image or a feature map, as a substitution of the standard grid sampling. Dai \etal~\cite{dai2017deformable} and Zhu \etal~\cite{zhu2019deformable} propose deformable convolutional networks, where the rectangular window sampling in standard convolution is replaced 
with shifted point sampling. Deformable DETR~\cite{zhu2021deformable} follows this manner and samples key points relative to a certain query to conduct deformable attention. Similar practices also take place when images are downsampled to low-res ones 
for content-aware image resizing, \textit{a.k.a.} seam carving~\cite{avidan2007seam}. \textit{E.g.}, Zhang \etal~\cite{zhang2019zoom} propose to learn to downsample an image 
with saliency guidance, in order to preserve more 
information of the original image, and Jin et al.~\cite{jin2022learning} also set a learnable deformation module to downsample the images.

Different from recent kernel-based upsamplers, we 
interpret the essence of upsampling as point re-sampling. Therefore in feature upsampling, we 
tend to follow the same spirit as the work above and use simple designs to 
achieve a strong and efficient dynamic upsampler.

\section{Learning to Sample and Upsample}

In this section we elaborate the designs of DySample and its variants. We first present a naive implementation and then show how to improve it step by step.

\subsection{Preliminary}
\begin{figure}[!t]
	\centering
	\includegraphics[width=\linewidth]{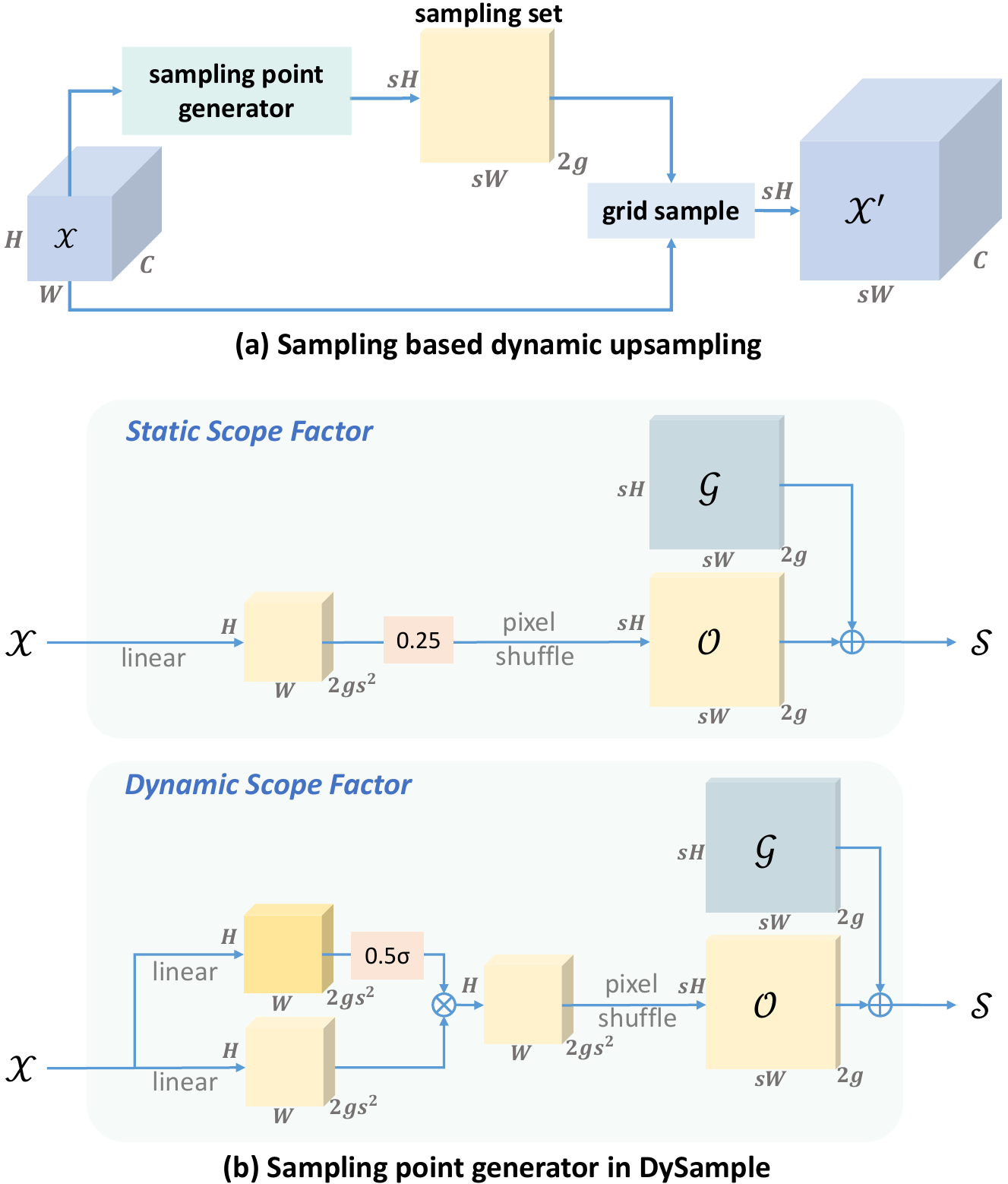}
	\caption{\textbf{Sampling based dynamic upsampling and module designs in DySample.} The input feature, upsampled feature, generated offset, and original grid are denoted by $\mathcal{X}$, $\mathcal{X}'$, $\mathcal{O}$, and $\mathcal{G}$, respectively. 
    (a) The sampling set is generated by the sampling point generator, with which the input feature is re-sampled by the $\tt{grid\_sample}$ function. In the generator (b), the sampling set is the sum of the generated offset and the original grid position. The upper 
    box shows the version with the `static scope factor', where the offset is generated with a linear layer. The bottom one 
    describes the version with `dynamic scope factor', where the a scope factor is first generated and then is used to 
    modulate the offset. `$\sigma$' denotes the $\tt{sigmoid}$ function.
    }
    \label{fig:pipeline}
\end{figure}

We return to the essence of upsampling, \ie, point sampling, in the light of modeling geometric information. With the built-in function in PyTorch, we first provide a naive implementation to demonstrate the feasibility of sampling based dynamic upsampling (Fig.~\ref{fig:pipeline}(a)).

\myparagraph{Grid Sampling.} Given a feature map $\mathcal{X}$ of size $C\times H_1 \times W_1$, and a sampling set $\mathcal{S}$ of size $2 \times H_2 \times W_2$, where $2$ of the first dimension denotes the $x$ and $y$ coordinates, the $\tt{grid\_sample}$ function uses the positions in $\mathcal{S}$ to re-sample the hypothetical bilinear-interpolated $\mathcal{X}$ into $\mathcal{X}'$ of size $C\times H_2 \times W_2$. 
This process is defined by
\begin{equation}
    \label{eq:sampling}
    \mathcal{X}'=\tt{grid\_sample}(\mathcal{X},\mathcal{S})\,.
\end{equation}

\myparagraph{Naive Implementation.} Given an upsampling scale factor of $s$ and a feature map $\mathcal{X}$ of size $C \times H \times W$, a linear layer, whose input and output channel numbers are $C$ and $2s^2$, is used to generate the offset $\mathcal{O}$ of size $2s^2 \times H \times W$, which is then reshaped to $2 \times sH \times sW$ by Pixel Shuffling~\cite{shi2016real}. Then the sampling set $\mathcal{S}$ is the sum of the offset $\mathcal{O}$ and the original sampling grid $\mathcal{G}$, \ie,
\begin{equation}
    \label{eq:offset}
    \mathcal{O}=\tt{linear}(\mathcal{X})\,,
\end{equation}
\begin{equation}
    \label{eq:sample_set}
    \mathcal{S}=\mathcal{G}+\mathcal{O}\,,
\end{equation}
where the reshaping operation is omitted. Finally the upsampled feature map $\mathcal{X}'$ of size $C \times sH \times sW$ can be generated with the sampling set by $\tt{grid\_sample}$ as Eq.~\eqref{eq:sampling}. 

This preliminary design 
obtains $37.9$ AP with Faster R-CNN~\cite{ren2015faster} on object detection~\cite{lin2014microsoft} and $41.9$ mIoU with SegFormer-B1~\cite{xie2021segformer} on semantic segmentation~\cite{zhou2017scene} (cf. CARAFE: $38.6$ AP and $42.8$ mIoU).
Next we present DySample upon this naive implementation. 

\subsection{DySample: Upsampling by Dynamic Sampling}
By studying the naive implementation, we observe that shared initial offset position among the $s^2$ upsampled points 
neglects the position relation, and that the unconstrained 
walking scope of offsets can cause 
disordered point sampling. We first discuss the two issues. We will also study implementation details such as feature groups and dynamic offset scope.

\begin{figure}[!t]
	\centering
	\includegraphics[width=\linewidth]{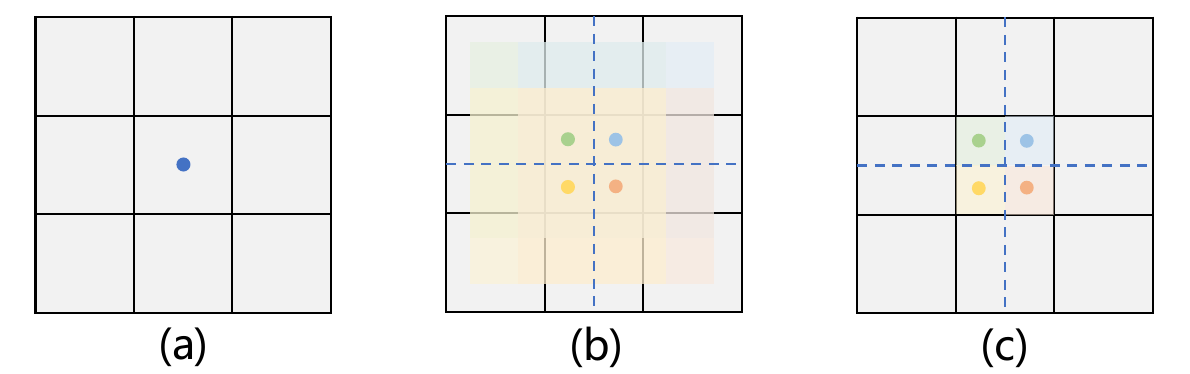}
	\caption{\textbf{Initial sampling positions and offset scopes.} The points and the colored masks represent the initial sampling positions and the offset scopes, respectively. Considering sampling four points ($s=2$), (a) in the case of nearest initialization, 
    the four offsets share the same initial position but ignore position relation; in bilinear initialization (b), we separate the initial positions such that they distribute evenly. 
    Without offset modulation (b), the offset scope would typically overlap, so in (c) we locally constrain the offset scope to reduce the overlap.
	}
	\label{fig:initpos}
\end{figure}

\myparagraph{Initial Sampling Position.} In the preliminary version, the $s^2$ sampling positions 
w.r.t. one point in $\mathcal{X}$ are all made 
fixed at the same initial position (the standard grid points in $\mathcal{X}$), as shown in Fig.~\ref{fig:initpos} (a). This practice ignores the position relation among the $s^2$ neighboring points 
such that the initial sampling positions distribute unevenly. If the generated offsets are all zeros, the upsampled feature is equivalent to the NN interpolated one. Hence, this preliminary initialization can be called `nearest initialization'. Targeting this problem, we alter the initial position to `bilinear initialization' as in Fig.~\ref{fig:initpos}(b), where zero offsets would bring the bilinearly interpolated feature map.


After changing the initial sampling position, the performance improves to $38.1$ ($\mycolor{+0.2}$) AP and $42.1$ ($\mycolor{+0.2}$) mIoU, as shown in Table~\ref{tab:ablation_initpos}.

\begin{figure}[!t]
	\centering
	\includegraphics[width=\linewidth]{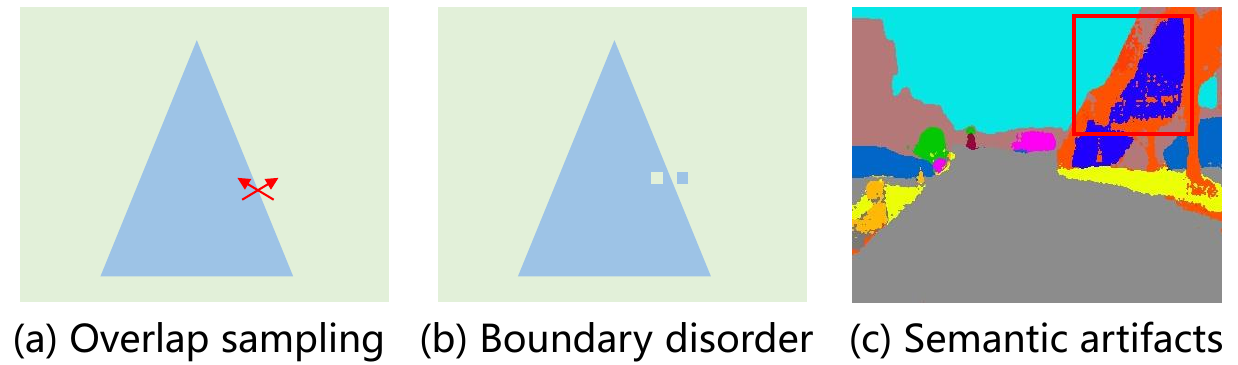}
	\caption{\textbf{Prediction artifacts due to offset overlap.} If the offsets overlap (a), the point value near boundaries may be in disorder (b), and the error would propagate layer by layer and finally cause prediction artifacts (c).
	}
	\label{fig:overlap}
\end{figure}

\begin{table}[!t] \footnotesize
    \centering
    \renewcommand{\arraystretch}{1.1}
    \addtolength{\tabcolsep}{0pt}
    \begin{tabular}{@{}lcc@{}}
    \toprule
        Sampling Initialization  & mIoU & $AP$ \\
        \midrule
        Nearest Initialization & 41.9 & 37.9 \\
        \rowcolor{LavenderBlush} Bilinear Initialization & \textbf{42.1} & \textbf{38.1} \\
    \bottomrule
    \end{tabular}
    \caption{Ablation study on initial sampling position.}
    \label{tab:ablation_initpos}
\end{table}

\myparagraph{Offset Scope.} Due to the existence of normalization layers, 
the values of one certain output feature are typically in the range of $[-1,1]$, centered at $0$. Therefore, the walking scope of the local $s^2$ sampling positions could 
overlap significantly, as shown in Fig.~\ref{fig:overlap}(a). The overlap would easily influence the prediction near boundaries (Fig.~\ref{fig:overlap}(b)), and such errors would 
propagate stage by stage and cause output 
artifacts (Fig.~\ref{fig:overlap}(c)). To alleviate this, we 
multiply the offset by a factor of $0.25$, which just meets the theoretical marginal condition between overlap and non-overlap. This factor is called the `static scope factor', such that the walking scope of the sampling positions is locally constrained, as shown in Fig.~\ref{fig:initpos}(c). Here we rewrite Eq.~\eqref{eq:offset} as
\begin{equation}
    \label{eq:offset0.25}
    \mathcal{O}=0.25~\tt{linear}(\mathcal{X})\,.
\end{equation}
By setting the scope factor to $0.25$, performance improves to $38.3$ ($\mycolor{+0.2}$) AP and $42.4$ ($\mycolor{+0.3}$) mIoU. We also test other possible factors, as shown in Table~\ref{tab:ablation_scope}.

\textit{Remark.} Multiplying the factor is a soft 
solution of 
the problem; it cannot completely solve it. We have also tried to strictly constrain the offset scope in $[-0.25,0.25]$ with $\tt{tanh}$ function, but it works worse. Perhaps 
the explicit constraint limits the representation power, \eg, the explicit constraint version cannot handle the situation where some certain position expects a shift lager than $0.25$.


\begin{figure}[!t]
\begin{minipage}{\linewidth}
\makeatletter\def\@captype{table}\makeatother
\begin{minipage}[c]{0.37\linewidth}\footnotesize
\vspace{5pt}
    \centering
    \renewcommand{\arraystretch}{1.1}
    \addtolength{\tabcolsep}{-1pt}
    \begin{tabular}{@{}ccc@{}}
    \toprule
        Factor & mIoU & $AP$ \\
        \midrule
        0.1 & 42.2 & 38.1 \\
        \rowcolor{LavenderBlush} 0.25 & \textbf{42.4} & \textbf{38.3} \\
        0.5 & 42.2 & 38.1 \\
        1 & 42.1 & 38.1 \\
    \bottomrule
    \end{tabular}
    \caption{Ablation study on the effect of static scope factor.}
    \label{tab:ablation_scope}
\end{minipage}
\hfill
\makeatletter\def\@captype{table}\makeatother
\begin{minipage}[c]{0.57\linewidth}\footnotesize
\vspace{-8pt}
    \centering
    \renewcommand{\arraystretch}{1.1}
    \addtolength{\tabcolsep}{-2pt}
    \begin{tabular}{@{}cccc@{}}
    \toprule
        Groups & Dynamic & mIoU & $AP$ \\
        \midrule
        1 &  & 42.4 & 38.3 \\
        1 & \checkmark & 42.6 & 38.4 \\
        4 &  & 43.2 & 38.6 \\
        \rowcolor{LavenderBlush} 4 & \checkmark & \textbf{43.3} & \textbf{38.7} \\
    \bottomrule
    \end{tabular}
    \caption{Ablation study on the effect of dynamic scope factor.}
    \label{tab:ablation_dynamic_scope}
\end{minipage}
\end{minipage}
\end{figure}

\begin{figure}[!t]
\centering
     \begin{subfigure}[b]{0.519\linewidth}
         \centering
         \includegraphics[width=\linewidth]{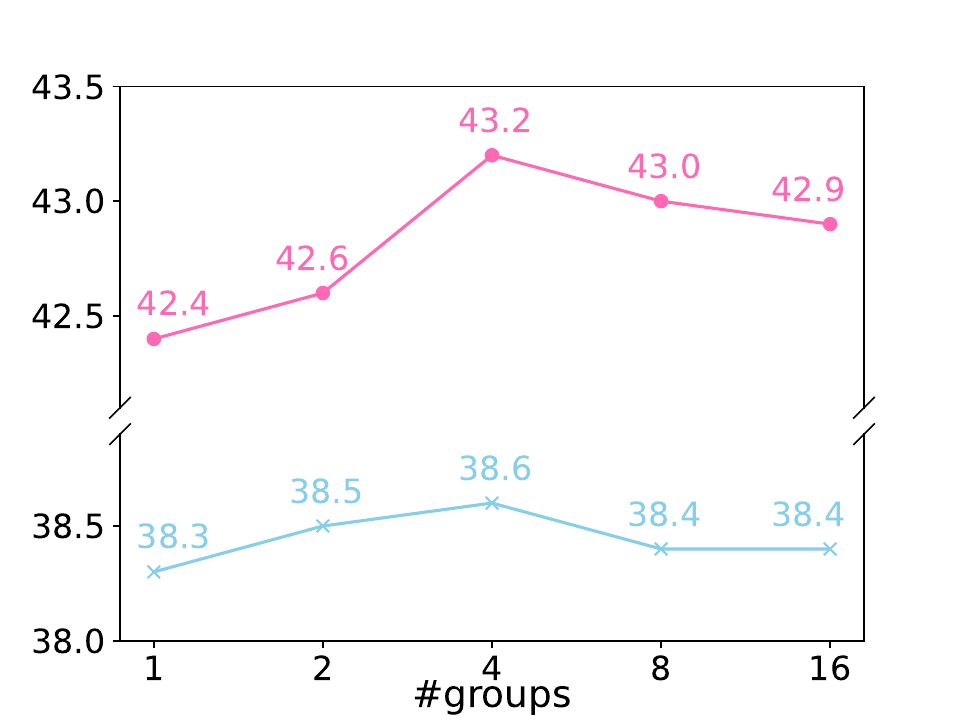}
         \caption{LP style}
     \end{subfigure}
     \hfill
     \begin{subfigure}[b]{0.471\linewidth}
         \centering
         \includegraphics[width=\linewidth]{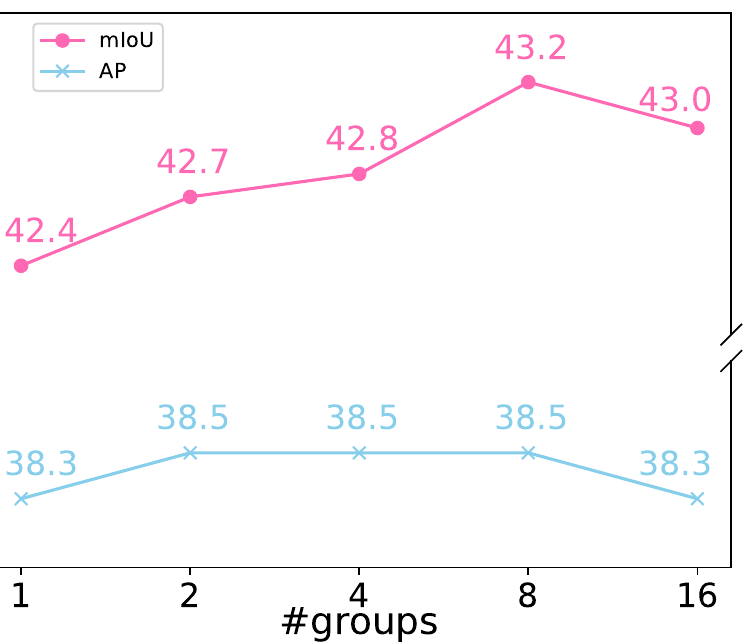}
         \caption{PL style}
     \end{subfigure}
    \caption{Ablation study on the number of feature groups.}
    \label{fig:ablation_group}
\end{figure}

\myparagraph{Grouping.} Here we study group-wise upsampling, where features share the same sampling set in each group. Specifically, one can divide the feature map into $g$ groups along the channel dimension and generate $g$ groups of offsets. 

According to Fig.~\ref{fig:ablation_group}, grouping works. When $g=4$, performance reaches to $38.6$ ($\mycolor{+0.3}$) AP and $43.2$ ($\mycolor{+0.8}$) mIoU.

\myparagraph{Dynamic Scope Factor.}
To increase the flexibility of the offset, we further generate point-wise `dynamic scope factors' by linear projecting the input feature. By using the $\tt{sigmoid}$ function and a $0.5$ static factor, the dynamic scope takes the value in the range of $[0,0.5]$, centered at $0.25$ as the static ones. The dynamic scope operation can refer to Fig.~\ref{fig:pipeline}(b). Here we rewrite Eq.~\eqref{eq:offset0.25} as
\begin{equation}
    \label{eq:offset0.5sigma}
    \mathcal{O}=0.5~\tt{sigmoid}(\tt{linear}_1(\mathcal{X}))\cdot\tt{linear}_2(\mathcal{X})\,.
\end{equation}
Per Table~\ref{tab:ablation_dynamic_scope}, the dynamic scope factor further boosts the performance to $38.7$ ($\mycolor{+0.1}$) AP and $43.3$ ($\mycolor{+0.1}$) mIoU. 

\begin{figure}[!t]
	\centering
	\includegraphics[width=\linewidth]{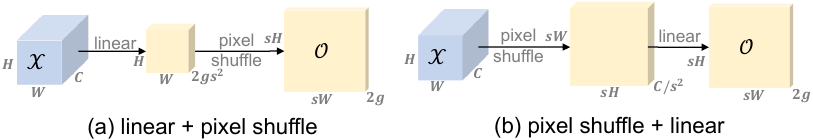}
	\caption{\textbf{Offset generation styles.} While the (a) `linear+pixel shuffle' (LP) version requires more parameters than the (b) `pixel shuffle + linear' (PL) version, the former is more flexible, consumes smaller memory footprint, and has faster inference speed.
	}
	\label{fig:style}
\end{figure}

\myparagraph{Offset Generation Styles.} In the design above, linear projection is first used to produce $s^2$ offset sets. The sets are then reshaped to satisfy the spatial size. We call this process as `linear+pixel shuffle' (LP). To save parameters and GFLOPs, we can execute the reshaping operation ahead, 
\ie, first reshaping the feature $\mathcal{X}$ to the size of $\frac{C}{s^2}\times sH \times sW$ and then linearly projecting it to $2g\times sH \times sW$. Similarly, we call this procedure `pixel shuffle+linear' (PL). With other hyper parameters fixed, the number of parameters can be reduced to $1/s^4$ under the PL setting. Through experiments, we empirically set the group number as $4$ and $8$ for the LP and PL version respectively according to Fig.~\ref{fig:ablation_group}. Further, we find that the PL version works better than the LP version on SegFormer (Table~\ref{tab:segformer}) and MaskFormer (Table~\ref{tab:maskformer}), but slightly worse on other tested models.

\myparagraph{DySample Series.} According to the form of scope factor (static/dynamic) and offset generation styles (LP/PL), we investigate four variants: 
\begin{itemize}
    \setlength\itemsep{-0.1em}
    \item[i)] DySample: LP-style with the static scope factor;

    \item[ii)] DySample+: LP-style with the dynamic scope factor;

    \item[iii)] DySample-S: PL-style with the static scope factor;

    \item[iv)] DySample-S+: PL-style with dynamic scope factor.
\end{itemize}

\subsection{How DySample works}
\begin{figure*}[!t]
	\centering
	\includegraphics[width=\linewidth]{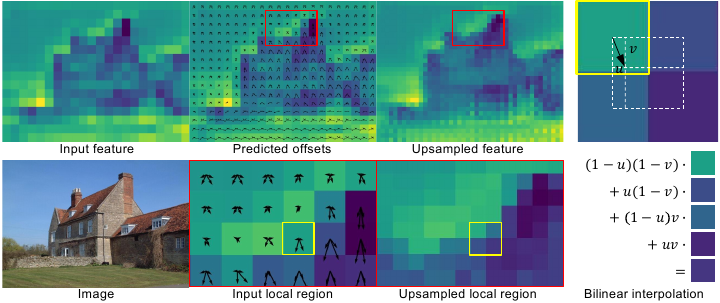}
	\caption{\textbf{Visualization of the upsampling process in DySample.} A part of the boundary in red box is highlighted for a close view. We generate content-aware offsets to construct new sampling points to resample the input feature map with bilinear interpolation. The new sampling positions are indicated by the arrowheads. The yellow boxed point in the low-res feature is selected to illustrate the bilinear interpolation process.}
	\label{fig:dysample_visual}
\end{figure*}
The sampling process of DySample is visualized in Fig.~\ref{fig:visual}. We highlight a (red boxed) local region to show how DySample divides one point on the edge to four to make the edge clearer. For the yellow boxed point, it generates four offsets pointing to the four upsampled points in sense of bilinear interpolation. In this example, the top left point is divided to the `sky' (lighter), while the other three are divided to the `house' (darker). The rightmost subplot indicates how the bottom right upsampled point is formed.

\subsection{Complexity Analysis}
\begin{figure}[!t]
	\centering
	\includegraphics[width=\linewidth]{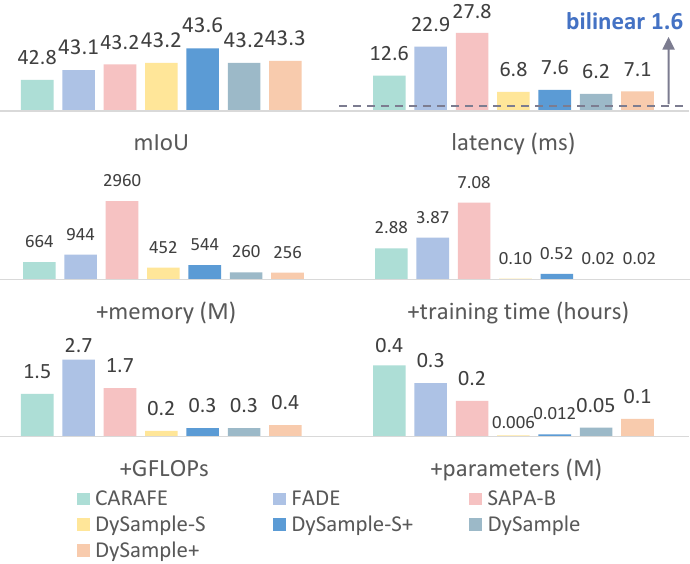}
	\caption{\textbf{Complexity analysis.} DySample series achieve the overall best performances on SegFormer-B1~\cite{xie2021segformer}, and cost the least latency, memory footprint, training time, GFLOPs, and number of parameters among the recent strong dynamic upsamplers. The inference time is tested by upsampling a $256\times 120\times 120$ feature map (and a $256\times 240\times 240$ guidance feature if needed) with a single Nvidia GTX 3090 GPU on a server. `+' means the additional amount compared with bilinear interpolation.}
	\label{fig:complexity}
\end{figure}
We use a random feature map of size $256\times 120\times 120$ (and a guidance map of size $256\times 240\times 240$ if required) as the input to test the inference latency. We use SegFormer-B1 to compare the performance, training memory, training time, GFLOPs, and number of parameters when bilinear interpolation (default) is replaced by other upsamplers. 

The quantitative results are shown in Fig.~\ref{fig:complexity}. Besides the best performances, DySample series cost the least inference latency, training memory, training time, GFLOPs, and number of parameters than all previous strong dynamic upsamplers. For the inference time, DySample series cost $6.2\sim7.6$ ms to upsample a $256\times 120\times 120$ feature map, which 
approaches to that of bilienar interpolation ($1.6$ms). 
Particularly, due to 
the use of highly optimized PyTorch built-in function, the backward propagation of DySample is 
rather fast; the increased training time is negligible. 

Among DySample series, the `-S' versions cost less parameters and GFLOPs, but more memory footprint and latency, because PL needs an extra storage of $\mathcal{X}$. The `+' versions also introduce a bit more computational amount.

\subsection{Discussion on Related Work}
Here we compare DySample with CARAFE~\cite{jiaqi2019carafe}, SAPA~\cite{lu2022sapa} and deformable attention~\cite{zhu2021deformable}.

\myparagraph{Relation to CARAFE.}
CARAFE generates content-aware upsampling kernels to reassemble the input feature. In DySample, we generate upsampling positions instead of kernels. Under the kernel-based view, DySample uses $2\times 2$ bilinear kernels, while CARAFE uses $5\times 5$ ones. In CARAFE if placing a kernel centered at a point, the kernel size must at least be $3\times 3$, so the GFLOPs is at least $2.25$ times larger than DySample. Besides, the upsampling kernel weights in 
CARAFE are learned, but in DySample they are conditioned on the $x$ and $y$ position. Therefore, to maintain a single kernel DySample only needs a 2-channel feature map (given that the group number $g=1$), but CARAFE requires a $K\times K$-channel one, which explains why DySample is more efficient. 

\myparagraph{Relation to SAPA.}
SAPA introduces the concept of semantic cluster into feature upsampling and views the upsampling process as finding a correct semantic cluster for each upsampling point. In DySample, offset generation can also be seen as seeking for a semantically 
similar region for each point. However, DySample does not need the guidance map and thus is more efficient and easy-to-use.

\myparagraph{Relation to Deformable Attention.}
Deformable attention~\cite{zhu2021deformable} mainly enhances features; it samples many points at each position to \textit{aggregate} them to form a new point. But DySample is tailored for upsampling; it samples a single point for each upsampled position to \textit{divide} one point to $s^2$ upsampled points. DySample reveals that sampling a single point for each upsampled position is enough as long as the upsampled $s^2$ points can be dynamically divided.

\section{Applications}

\begin{figure*}[!t]
	\centering
	\includegraphics[width=\linewidth]{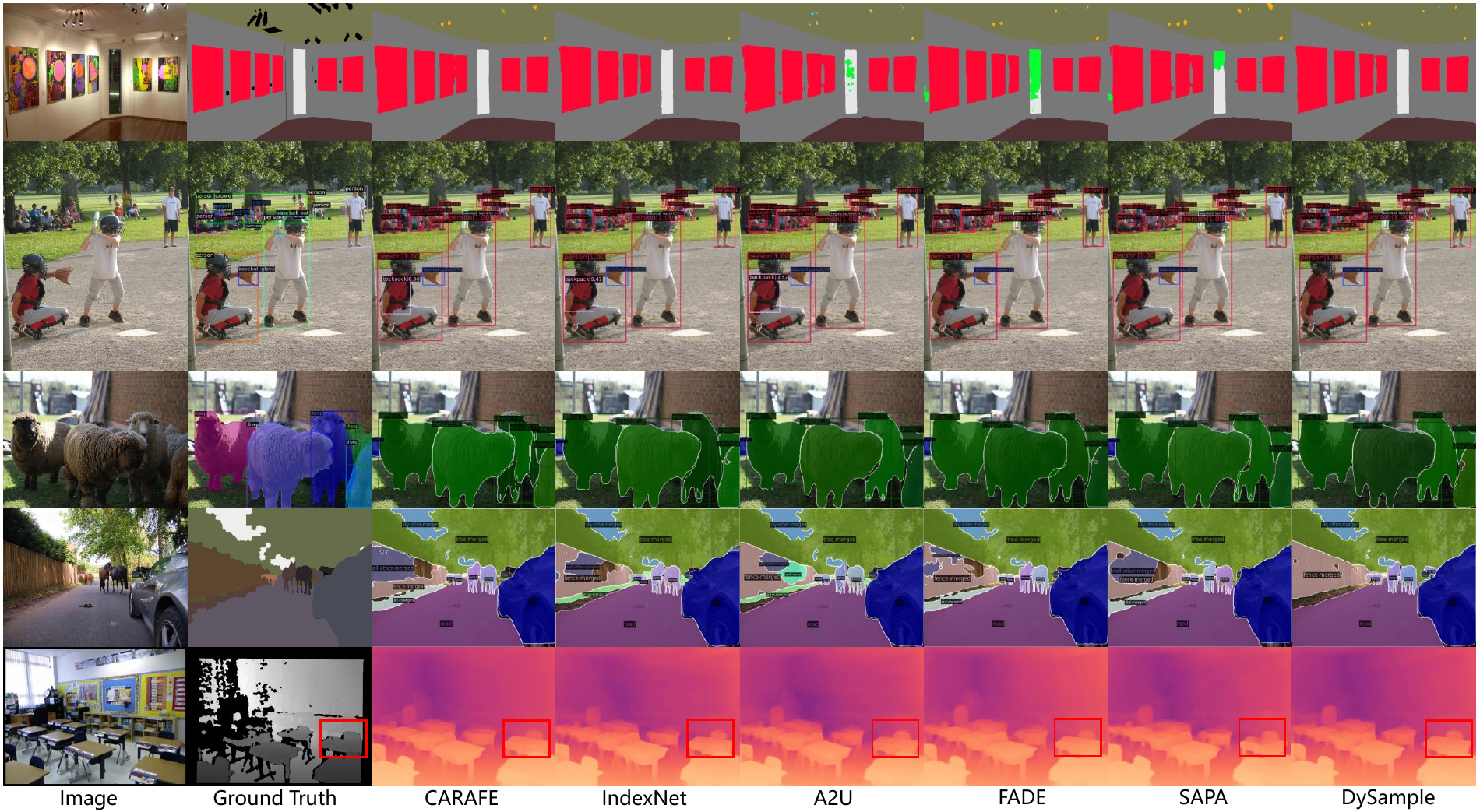}\vspace{-5pt}
	\caption{\textbf{Qualitative visualizations.} From top to bottom: semantic segmentation, object detection, instance segmentation, panoptic segmentation, and monocular depth estimation.
	}
	\label{fig:visual}
\end{figure*}

Here we apply DySample on five dense prediction tasks, including 
semantic segmentation, 
object detection, 
instance segmentation, 
panoptic segmentation, and 
depth estimation. Among the upsampler competitors, in bilinear interpolation, we set the scale factor as $2$ and `align corners' as False. For deconvolution, we set the kernel size as $3$, the stride as $2$, the padding as $1$ and the output padding as $1$. For pixel shuffle~\cite{shi2016real}, we first use a $3$-kernel size convolution to increase the channel number to $4$ times of the original one, and then apply the `pixel shuffle' function. For CARAFE~\cite{jiaqi2019carafe}, we adopt its default setting. The `HIN' version of IndexNet~\cite{lu2019indices} and the `dynamic-cs-d\dag' version of A2U~\cite{dai2021learning} are used. FADE~\cite{lu2022fade} without gating mechanism and SAPA-B~\cite{lu2022sapa} are used because they are more stable across all the dense prediction tasks. 

\subsection{Semantic Segmentation}

Semantic segmentation infers per-pixel class labels. 
Upsamplers are 
often adopted several times to obtain the high-res output in typical models. The precise per-pixel prediction is largely dependent on the upsampling quality.

\myparagraph{Experimental Protocols.}
We use the 
ADE20K~\cite{zhou2017scene} 
data set. Besides the commonly used mIoU metric, we also report the bIoU~\cite{cheng2021boundary} metric to evaluate the boundary quality. 
We first use a light-weight 
baseline SegFormer-B1~\cite{xie2021segformer}, 
where $3+2+1=6$ upsampling stages are involved, and then 
test DySample on a stronger baseline MaskFormer~\cite{cheng2021per}, with Swin-B~\cite{liu2021swin} and Swin-L as the backbone, where $3$ upsampling stages are involved in FPN. We use the official codebase provided by the authors and follow all the training settings except for only modifying the upsampling stages.

\begin{table}[!t]\footnotesize
    \centering
    \renewcommand{\arraystretch}{1.1}
    \addtolength{\tabcolsep}{0pt}
    \begin{tabular}{@{}lllcc@{}}
    \toprule
        SegFormer-B1 & FLOPs & Params & mIoU & bIoU \\
        \midrule
        Bilinear & 15.9 & 13.7M & 41.68 & 27.80 \\
        Deconv & +34.4 & +3.5M & 40.71 & 25.94 \\
        PixelShuffle~\cite{shi2016real} & +34.4 & +14.2M & 41.50 & 26.58 \\
        CARAFE~\cite{jiaqi2019carafe} & +1.5 & +0.4M & 42.82 & 29.84 \\
        \rowcolor{WhiteSmoke} IndexNet~\cite{lu2019indices} & +30.7 & +12.6M & 41.50 & 28.27 \\
        \rowcolor{WhiteSmoke} A2U~\cite{dai2021learning} & +0.4 & +0.1M & 41.45 & 27.31 \\
        \rowcolor{WhiteSmoke} FADE~\cite{lu2022fade} & +2.7 & +0.3M & 43.06 & \textbf{31.68} \\
        \rowcolor{WhiteSmoke} SAPA-B~\cite{lu2022sapa} & +1.0 & +0.1M & 43.20 & \underline{30.96} \\
        DySample-S & +0.2 & +6.1K & 43.23 & 29.53 \\
        DySample-S+ & +0.3 & +12.3K & \textbf{43.58} & 29.93 \\
        DySample & +0.3 & +49.2K & 43.21 & 29.12 \\
        DySample+ & +0.4 & +0.1M & \underline{43.28} & 29.23 \\
    \bottomrule
    \end{tabular}
    \caption{Semantic segmentation results with SegFormer-B1 on ADE20K. Best performance is in \textbf{boldface} and second best is \underline{underlined}.}
    \label{tab:segformer}
\end{table}

\begin{table}[!t]\footnotesize
    \centering
    \renewcommand{\arraystretch}{1.1}
    \addtolength{\tabcolsep}{3pt}
    \begin{tabular}{@{}clc@{}}
    \toprule
        Backbone & Upsampler & mIoU \\
        \midrule
        Swin-B & Nearest & 52.70 \\
        & CARAFE & \underline{53.53} \\
        & DySample-S+ & \textbf{53.91}\\
        Swin-L & Nearest & 54.10 \\
        & CARAFE & \underline{54.61} \\
        & DySample-S+ & \textbf{54.90}\\
    \bottomrule
    \end{tabular}
    \caption{Semantic segmentation results with MaskFormer on ADE20K. Best performance is in \textbf{boldface} and second best is \underline{underlined}.}
    \label{tab:maskformer}
\end{table}

\myparagraph{Semantic Segmentation Results.}
Quantitative results are shown in Tables~\ref{tab:segformer} and~\ref{tab:maskformer}. We can see DySample achieves the best mIoU metric of $43.58$ on SegFormer-B1, but the bIoU metric is lower than those 
guided upsamplers such as FADE and SAPA. Therefore we can infer that DySample improves the performance mainly from the interior regions, and the guided upsamplers mainly 
improve boundary quality. As shown in Fig.~\ref{fig:visual} row 1, the output of DySample is similar to that of CARAFE, but more distinctive near boundaries; the guided upsamplers predict sharper boundaries, but 
have wrong predictions on interior regions. For the stronger baseline MaskFormer, DySample also improves the mIoU metric from $52.70$ to $53.91$ ($\mycolor{+1.21}$) with Swin-B and from $54.10$ to $54.90$ ($\mycolor{+0.80}$) with Swin-L.

\subsection{Object Detection and Instance Segmentation}
\begin{table}[!t] \footnotesize
\centering
\renewcommand{\arraystretch}{1.1}
\addtolength{\tabcolsep}{-4.9pt}
\begin{tabular}{@{}lclcccccc@{}}
\toprule
Faster R-CNN & Backbone & Params & $AP$ & $AP_{50}$ & $AP_{75}$ & $AP_S$  & $AP_M$  & $AP_{L}$  \\
\midrule
Nearest & R50 & 46.8M & 37.5 & 58.2 & 40.8 & 21.3 & 41.1 & 48.9 \\
Deconv & R50 & +2.4M & 37.3 & 57.8 & 40.3 & 21.3 & 41.1 & 48.0 \\
PixelShuffle~\cite{shi2016real} & R50 & +9.4M & 37.5 & 58.5 & 40.4 & 21.5 & 41.5 & 48.3 \\
CARAFE~\cite{jiaqi2019carafe} & R50 & +0.3M & \underline{38.6} & \underline{59.9} & \textbf{42.2} & \textbf{23.3} & \underline{42.2} & 49.7 \\
\rowcolor{WhiteSmoke} IndexNet~\cite{lu2019indices} & R50 & +8.4M & 37.6 & 58.4 & 40.9 & 21.5 & 41.3 & 49.2 \\
\rowcolor{WhiteSmoke} A2U~\cite{dai2021learning} & R50 & +38.9K & 37.3 & 58.7 & 40.0 & 21.7 & 41.1 & 48.5 \\
\rowcolor{WhiteSmoke} FADE~\cite{lu2022fade} & R50 & +0.2M & 38.5 & 59.6 & 41.8 & \underline{23.1} & \underline{42.2} & 49.3 \\
\rowcolor{WhiteSmoke} SAPA-B~\cite{lu2022sapa} & R50 & +0.1M & 37.8 & 59.2 & 40.6 & 22.4 & 41.4 & 49.1 \\
DySample-S & R50 & +4.1K & 38.5 & 59.5 & \underline{42.1} & 22.7 & 41.9 & \textbf{50.2} \\
DySample-S+ & R50 & +8.2K & \underline{38.6} & 59.8 & \underline{42.1} & 22.5 & 42.1 & \underline{50.0} \\
DySample & R50 & +32.7K & \underline{38.6} & \underline{59.9} & 42.0 & 22.9 & 42.1 & \textbf{50.2} \\
DySample+ & R50 & +65.5K & \textbf{38.7} & \textbf{60.0} & \textbf{42.2} & 22.5 & \textbf{42.4} & \textbf{50.2} \\
\midrule
Nearest & R101 & 65.8M & 39.4 & 60.1 & 43.1 & 22.4 & 43.7 & 51.1 \\
DySample+ & R101 & +65.5K & \textbf{40.5} & \textbf{61.6} & \textbf{43.8} & \textbf{24.2} & \textbf{44.5} & \textbf{52.3} \\
\bottomrule
\end{tabular}
\caption{Object detection results with Faster R-CNN on MS COCO. Best performance is in \textbf{boldface} and second best is \underline{underlined}.}
\label{tab:faster_rcnn}
\end{table}

\begin{table}[!t] \footnotesize
\centering
\renewcommand{\arraystretch}{1.111}
\addtolength{\tabcolsep}{-4.5pt}
\begin{tabular}{@{}lcccccccc@{}}
\toprule
Mask R-CNN & Task & Backbone & $AP$ & $AP_{50}$ & $AP_{75}$ & $AP_S$ & $AP_M$ & $AP_{L}$ \\
\midrule
Nearest  & Bbox & R50 & 38.3 & 58.7 & 42.0 & 21.9 & 41.8 & 50.2 \\
Deconv  &  & R50 & 37.9 & 58.5 & 41.0 & 22.0 & 41.6 & 49.0 \\
PixelShuffle~\cite{shi2016real}  &  & R50 & 38.5 & 59.4 & 41.9 & 22.0 & 42.3 & 49.8 \\
CARAFE~\cite{jiaqi2019carafe} & & R50 & 39.2 & 60.0 & \underline{43.0} & 23.0 & \underline{42.8} & 50.8 \\
\rowcolor{WhiteSmoke} IndexNet~\cite{lu2019indices} & & R50 & 38.4 & 59.2 & 41.7 & 22.1 & 41.7 & 50.3 \\
\rowcolor{WhiteSmoke} A2U~\cite{dai2021learning} & & R50 & 38.2 & 59.2 & 41.4 & 22.3 & 41.7 & 49.6 \\
\rowcolor{WhiteSmoke} FADE~\cite{lu2022fade} & & R50 & 39.1 & \underline{60.3} & 42.4 & \textbf{23.6} & 42.3 & 51.0 \\
\rowcolor{WhiteSmoke} SAPA-B~\cite{lu2022sapa} & & R50 & 38.7 & 59.7 & 42.2 & 23.1 & 41.8 & 49.9\\
DySample-S & & R50 & \underline{39.3} & \textbf{60.4} & \underline{43.0} & 23.2 & 42.7 & \underline{51.1}\\
DySample-S+ & & R50 & \underline{39.3} & \underline{60.3} & 42.8 & 23.2 & 42.7 & 50.8\\
DySample & & R50 & 39.2 & \underline{60.3} & \underline{43.0} & \underline{23.5} & 42.5 & 51.0\\
DySample+ & & R50 & \textbf{39.6} & \textbf{60.4} & \textbf{43.5} & 23.4 & \textbf{42.9} & \textbf{51.7}\\
\midrule
Nearest & & R101 & 40.0 & 60.4 & 43.7 & 22.8 & 43.7 & 52.0 \\
DySample+ & & R101 & \textbf{41.0} & \textbf{61.9} & \textbf{44.9} & \textbf{24.3} & \textbf{45.0} & \textbf{53.5}\\
\midrule
Nearest & Segm & R50 & 34.7 & 55.8 & 37.2 & 16.1 & 37.3 & 50.8 \\
Deconv  & & R50 & 34.5 & 55.5 & 36.8 & 16.4 & 37.0 & 49.5 \\
PixelShuffle~\cite{shi2016real} & & R50 & 34.8 & 56.0 & 37.3 & 16.3 & 37.5 & 50.4 \\
CARAFE~\cite{jiaqi2019carafe} & & R50 & 35.4 & 56.7 & 37.6 & 16.9 & \underline{38.1} & 51.3 \\
\rowcolor{WhiteSmoke} IndexNet~\cite{lu2019indices} & & R50 & 34.7 & 55.9 & 37.1 & 16.0 & 37.0 & 51.1 \\
\rowcolor{WhiteSmoke} A2U~\cite{dai2021learning} & & R50 & 34.6 & 56.0 & 36.8 & 16.1 & 37.4 & 50.3 \\
\rowcolor{WhiteSmoke} FADE~\cite{lu2022fade} & & R50 & 35.1 & 56.7 & 37.2 & 16.7 & 37.5 & 51.4 \\
\rowcolor{WhiteSmoke} SAPA-B~\cite{lu2022sapa} & & R50 & 35.1 & 56.5 & 37.4 & 16.7 & 37.6 & 50.6\\
DySample-S & & R50 & 35.4 & 56.8 & \underline{37.8} & 16.7 & 38.0 & 51.4\\
DySample-S+ & & R50 & \underline{35.5} & 56.8 & \underline{37.8} & 17.0 & 37.9 & \textbf{51.9}\\
DySample & & R50 & 35.4 & \underline{56.9} & \underline{37.8} & \underline{17.1} & 37.7 & 51.1\\
DySample+ & & R50 & \textbf{35.7} & \textbf{57.3} & \textbf{38.2} & \textbf{17.3} & \textbf{38.2} & \underline{51.8}\\
\midrule
Nearest & & R101 & 36.0 & 57.6 & 38.5 & 16.5 & 39.3 & 52.2 \\
DySample+ & & R101 & \textbf{36.8} & \textbf{58.7} & \textbf{39.5} & \textbf{17.5} & \textbf{40.0} & \textbf{53.8}\\
\bottomrule
\end{tabular}
\caption{Instance segmentation results with Mask R-CNN on MS COCO. The parameter increment is identical as in Faster R-CNN. Best performance is in \textbf{boldface} and second best is \underline{underlined}.}
\label{tab:mask_rcnn}
\end{table}

Being instance-level tasks, object detection aims to localize and classify objects, while instance segmentation need to further segment the objects. The quality of the upsampled features can have large effect on the classification, localization, and segmentation accuracy. 
\myparagraph{Experimental Protocols.}
We use the MS COCO~\cite{lin2014microsoft} data set. 
The $AP$ series metrics are reported. 
Faster R-CNN~\cite{ren2015faster} and Mask R-CNN~\cite{he2017mask} are chosen as the baselines. 
We modify the upsamplers in the FPN architecture for performance comparison. There are four and three upsampling stages in the FPN of Faster R-CNN and of Mask R-CNN, respectively. We use the code provided by \texttt{mmdetection}~\cite{chen2019mmdetection} and follow the $1\times$ training settings.

\myparagraph{Object Detection and Instance Segmentation Results.}
Quantitative results are shown in Tables~\ref{tab:faster_rcnn} and~\ref{tab:mask_rcnn}. Results show that DySample outperforms all 
compared upsamplers. With R50, DySample achieves the best performance among all tested upsamplers. When a stronger backbone is used, 
notable improvements can also be witnessed (R50 $+1.2$ vs.\ R101 $+1.1$ box AP on Faster R-CNN, and R50 $+1.0$ vs.\ R101 $+0.8$ mask AP on Mask R-CNN).

\subsection{Panoptic Segmentation}
\begin{table}[!t] \footnotesize
\centering
\renewcommand{\arraystretch}{1.1}
\addtolength{\tabcolsep}{-3pt}
\begin{tabular}{@{}lclccccc@{}}
\toprule
Panoptic FPN & Backbone & Params & $PQ$ & $PQ^{th}$ & $PQ^{st}$ & $SQ$ & $RQ$ \\
\midrule 
Nearest & R50 & 46.0M & 40.2 & 47.8 & 28.9 & 77.8 & 49.3 \\
Deconv & R50 & +1.8M & 39.6 & 47.0 & 28.4 & 77.1 & 48.5 \\
PixelShuffle~\cite{shi2016real} & R50 & +7.1M & 40.0 & 47.4 & 28.8 & 77.1 & 49.1 \\
CARAFE~\cite{jiaqi2019carafe} & R50 & +0.2M & 40.8 & 47.7 & 30.4 & 78.2 & 50.0 \\
\rowcolor{WhiteSmoke} IndexNet~\cite{lu2019indices} & R50 & +6.3M & 40.2 & 47.6 & 28.9 & 77.1 & 49.3 \\
\rowcolor{WhiteSmoke} A2U~\cite{dai2021learning} & R50 & +29.2K & 40.1 & 47.6 & 28.7 & 77.3 & 48.0 \\
\rowcolor{WhiteSmoke} FADE~\cite{lu2022fade} & R50 & +0.1M & 40.9 & 48.0 & 30.3 & 78.1 & 50.1 \\
\rowcolor{WhiteSmoke} SAPA-B~\cite{lu2022sapa} & R50 & +0.1M & 40.6 & 47.7 & 29.8 & 78.0 & 49.6 \\
DySample-S & R50 & +3.1K & 40.6 & 48.0 & 29.6 & 78.0 & 49.8 \\
DySample-S+ & R50 & +6.2K & 41.1 & \underline{48.1} & 30.5 & 78.2 & \underline{50.2} \\
DySample & R50 & +24.6K & \underline{41.4} & \textbf{48.5} & \underline{30.7} & \textbf{78.6} & \textbf{50.7} \\
DySample+ & R50 & +49.2K & \textbf{41.5} & \textbf{48.5} & \textbf{30.8} & \underline{78.3} & \textbf{50.7} \\
\midrule
Nearest & R101 & 65.0M & 42.2 & 50.1 & 30.3 & 78.3 & 51.4 \\
DySample+ & R101 & +49.2K & \textbf{43.0} & \textbf{50.2} & \textbf{32.1} & \textbf{78.6} & \textbf{52.4}\\
\bottomrule
\end{tabular}
\caption{Panoptic segmentation results with Panoptic FPN on MS COCO. Best performance is in \textbf{boldface} and second best is \underline{underlined}.}
\label{tab:panoptic_fpn}
\end{table}


\begin{table*}[!t]
    \centering
    \renewcommand{\arraystretch}{1.1}
    \addtolength{\tabcolsep}{1.5pt}
    \begin{tabular}{@{}l c ccc ccccc@{}}
    \toprule
        DepthFormer & Params & $\delta < 1.25$ & $\delta < 1.25^2$ & $\delta < 1.25^3$ & Abs Rel & RMS & log10 & RMS(log) & Sq Rel \\
        \midrule
        Bilinear & 47.6M & 0.873 & 0.978 & \underline{0.994} & 0.120 & 0.402 & \underline{0.050} & 0.148 & 0.071\\
        Deconv & +7.1M & 0.872 & \textbf{0.980} & \textbf{0.995} & \underline{0.117} & 0.401 & \underline{0.050} & 0.147 & \textbf{0.067}\\
        PixelShuffle & +28.2M & 0.874 & \underline{0.979} & \textbf{0.995} & \underline{0.117} & \underline{0.395} & \textbf{0.049} & \underline{0.146} & \underline{0.068}\\
        CARAFE~\cite{jiaqi2019carafe} & +0.3M & \underline{0.877} & 0.978 & \textbf{0.995} & \textbf{0.116} & 0.397 & \textbf{0.049} & \underline{0.146} & 0.069\\
        \rowcolor{WhiteSmoke} IndexNet~\cite{lu2019indices} & +6.3M & 0.873 & \textbf{0.980} & \textbf{0.995} & \underline{0.117} & 0.401 & \textbf{0.049} & 0.147 & \textbf{0.067}\\
        \rowcolor{WhiteSmoke} A2U~\cite{dai2021learning} & +30.0K & 0.874 & \underline{0.979} & \textbf{0.995} & 0.118 & 0.397 & \textbf{0.049} & 0.147 & \underline{0.068}\\
        \rowcolor{WhiteSmoke} FADE~\cite{lu2022fade} & +0.2M & 0.874 & 0.978 & \underline{0.994} & 0.118 & 0.399 & \textbf{0.049} & 0.147 & 0.071 \\
        \rowcolor{WhiteSmoke} SAPA-B & +0.1M & 0.870 & 0.978 & \textbf{0.995} & \underline{0.117} & 0.406 & \underline{0.050} & 0.149 & 0.069\\
        DySample-S & +5.8K & 0.871 & \underline{0.979} & \textbf{0.995} & 0.118 & 0.402 & \underline{0.050} & 0.148 & 0.069\\
        DySample-S+ & +11.5K & 0.872 & 0.978 & \underline{0.994} & 0.119 & 0.398 & \underline{0.050} & 0.148 & 0.070\\
        DySample & +46.1K & 0.872 & \underline{0.979} & \textbf{0.995} & \underline{0.117} & 0.400 & \underline{0.050} & 0.147 & \underline{0.068}\\
        DySample+ & +92.2K & \textbf{0.878} & \textbf{0.980} & \textbf{0.995} & \textbf{0.116} & \textbf{0.393} & \textbf{0.049} & \textbf{0.145} & \underline{0.068}\\
    \bottomrule
    \end{tabular}
    \caption{Monocular depth estimation results with DepthFormer (Swin-T) on NYU Depth V2. Best performance is in \textbf{boldface} and second best is \underline{underlined}.}
    \label{tab:depth_estimation}
\end{table*}

Panoptic segmentation 
is the joint task of 
semantic segmentation and instance segmentation. In this context, the upsamplers face the difficulty to discriminate instance boundaries, 
which places high demands on good semantic perception and discriminative ability of the upsamplers. 

\myparagraph{Experimental Protocols.}
We also conduct experiments on the MS COCO~\cite{lin2014microsoft} data set and report the PQ, SQ, and RQ metrics~\cite{kirillov2019panoptic}. 
We adopt Panoptic FPN~\cite{kirillov2019panopticfpn} as the baseline and \texttt{mmdetection} as our codebase. The default training setting is used to ensure a fair comparison. We only modify the total three upsampling stages in FPN.

\myparagraph{Panoptic Segmentation Results.} The quantitative results shown in Table~\ref{tab:panoptic_fpn} demonstrate that DySample 
invites consistent performance gains, \ie, $1.2$ and $0.8$ PQ improvement for R50 and R101 backbone respectively.

\subsection{Monocular Depth Estimation}
Monocular depth estimation requires a model to 
estimate a per-pixel depth map from a single image. A 
high-quality upsampler for depth estimation should simultaneously recover the details, maintain the consistency of the depth value in a plain region, and also tackles gradually changed depth values. 
\myparagraph{Experimental Protocols.}
We conduct the experiments on the NYU Depth V2 data set~\cite{silberman2012indoor} and report the $\delta <1.25$, $\delta <1.25^2$ and $\delta <1.25^3$ accuracy, absolute relative error (Abs Rel), root mean squared error (RMS) and its log version (RMS(log)), average log10 error (log10), and squared relative error (Sq Rel). We adopt DepthFormer-SwinT~\cite{li2022depthformer} as the baseline including four upsampling stages in the fusion module. For reproducibility, we use the codebase provided by \texttt{monocular depth estimation tool box}~\cite{lidepthtoolbox2022} and follow its recommended training settings, while only modifying the upsamplers.
\myparagraph{Monocular Depth Estimation Results.}
Quantitative results are shown in Table~\ref{tab:depth_estimation}. Among all upsamplers, DySample+ achieves the best performance, with an increase of $0.05$ in $\delta < 1.25$ accuracy, a decrease of $0.04$ in Abs Rel, and a decrease of $0.09$ in RMS compared with bilinear upsampling. Further, the qualitative comparison in Fig.~\ref{fig:visual} row $5$ also verifies the superiority of DySample, \eg, 
the accurate, consistent depth map of the chair. 

\section{Conclusion}
We propose DySample, a fast, effective, and universal dynamic upsampler. Different from common kernel based dynamic upsampling, DySample is designed from the perspective of point sampling. We start from a naive design and show how to gradually improve its performance from our deep insight of upsampling. 
Compared with other dynamic upsamplers, DySample not only reports the best performance but also gets rid of customised CUDA packages and consumes the least computational resources, showing superiority across latency, training memory, training time, GFLOPs, and number of parameters. For future work, we plan to 
apply DySample to 
low-level tasks and study joint modeling of upsampling and downsampling.

\vspace{5pt}
\noindent\textbf{Acknowledgement.} This work is supported by the National Natural Science Foundation of China under Grant No. 62106080.

{\small
\bibliographystyle{ieee_fullname}
\bibliography{egbib}
}

\end{document}